\documentclass[letterpaper, 10 pt, conference]{ieeeconf}  

\IEEEoverridecommandlockouts                              

\usepackage{graphicx} 
\usepackage{amsmath} 
\usepackage{amssymb}  
\let\labelindent\relax 
\usepackage{enumitem}
\usepackage{enumitem}
\usepackage{algorithm}
\usepackage{algorithmic}

\usepackage{booktabs}
\usepackage{multirow}
\usepackage{hyperref}

\usepackage{type1cm} 
\usepackage[T1]{fontenc}
\usepackage{textcomp}

\title{\LARGE \bf
A Hybrid Force-Position Strategy for Shape Control of Deformable Linear Objects With Graph Attention Networks
}

\author{Yanzhao Yu$^{*}$, Haotian Yang$^{*}$, Junbo Tan$^{\dagger}$, and Xueqian Wang$^{\dagger}$
\thanks{$^*$Equal contribution.}
\thanks{$^\dagger$Corresponding author.}
\thanks{The authors are with the Center of Intelligent Control and Telescience, Tsinghua Shenzhen International Graduate School, Tsinghua University, Shenzhen 518055, China (E-mail: \url{yuyz24@mails.tsinghua.edu.cn}, \url{yang-ht22@mails.tsinghua.edu.cn}, \url{tjblql@sz.tsinghua.edu.cn}, \url{wang.xq@sz.tsinghua.edu.cn}).}
\thanks{This work was supported by the Natural Science Foundation of Shenzhen (No.JCYJ20230807111604008, No. JCYJ20240813112007010), the Natural Science Foundation of Guangdong Province (No.2024A1515010003) and National Key Research and Development Program of China (No. 2022YFB4701400).}
}

\begin{document}

\maketitle
\thispagestyle{empty}
\pagestyle{empty}

\begin{abstract}

Manipulating deformable linear objects (DLOs) such as wires and cables is crucial in various applications like electronics assembly and medical surgeries. However, it faces challenges due to DLOs' infinite degrees of freedom, complex nonlinear dynamics, and the underactuated nature of the system. To address these issues, this paper proposes a hybrid force-position strategy for DLO shape control. The framework, combining both force and position representations of DLO, integrates state trajectory planning in the force space and Model Predictive Control (MPC) in the position space. We present a dynamics model with an explicit action encoder, a property extractor and a graph processor based on Graph Attention Networks. The model is used in the MPC to enhance prediction accuracy. Results from both simulations and real-world experiments demonstrate the effectiveness of our approach in achieving efficient and stable shape control of DLOs. Codes and videos are available at \href{https://sites.google.com/view/dlom}{https://sites.google.com/view/dlom}.

\end{abstract}

\section{INTRODUCTION}

Deformable linear objects (DLOs), such as wires, cables, and soft tubes, are widely encountered in practical applications like electronics assembly, medical guide wire placement, and crafts such as bamboo weaving. In these scenarios, precise shape control of DLOs is crucial. Automating this task with robots can significantly enhance efficiency.

However, several challenges complicate DLO shape control. First, DLOs possess infinite degrees of freedom, making their state difficult to represent in a low-dimensional vector with sufficient accuracy. Second, it is challenging to build accurate dynamics models of DLOs. DLOs exhibit highly nonlinear and dynamic behavior that depends on their material properties, applied forces, and deformation patterns. Third, due to the underactuated nature of the DLO system (i.e., the robot has fewer degrees of freedom than the object it manipulates), the DLO manipulation often suffers from instability and inefficiency, especially during large deformations.

\begin{figure*}[htbp]
    \centering
    \includegraphics[width=0.95\textwidth]{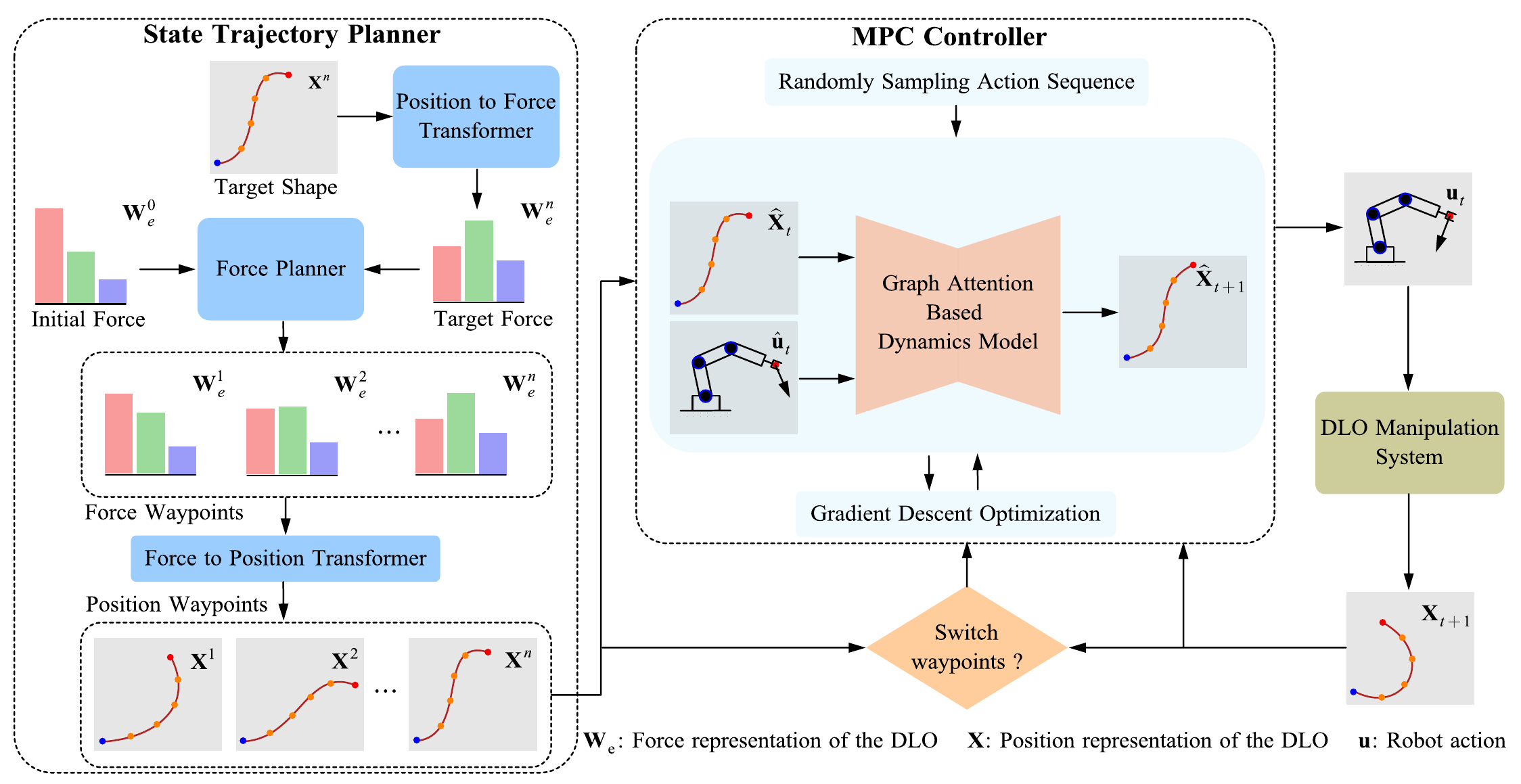}
    \caption{Overview of the proposed hybrid force-position strategy for DLO shape control. The whole framework includes a state trajectory planner and a MPC controller. The state trajectory planner first plans the DLO's trajectory in the force space, followed by conversion to the position space. The position waypoints are then used as intermediate goals for the MPC controller, which employs a GAT-based dynamics model to predict the DLO's future state.}
    \label{fig:overview-strategy}
    \vspace{-10pt}
\end{figure*}

To address the above challenges, this paper proposes a hybrid force-position strategy for the manipulation of DLOs, integrating both force and position information to achieve more effective shape control, as shown in Fig. \ref{fig:overview-strategy}. The whole framework consists of a state trajectory planner and a controller based on Model Predictive Control (MPC). The state trajectory of the DLO is first planned in the force space, then converted to the position space. The position waypoints are sequentially passed as intermediate goals to the MPC controller to optimize the robot's action sequence. Specifically, the MPC controller uses a dynamics model based on Graph Attention Networks (GAT) to predict the future state of the DLO and optimize the action sequence through gradient descent method. We conduct detailed simulation and real-world experiments to demonstrate the superiority of the proposed dynamics model and the effectiveness of the hybrid force-position strategy. 
The key contributions of this paper are highlighted as follows:
\begin{enumerate}[label=\arabic*)]
    \item We propose a hybrid state representation scheme combining position representation based on graph structures and force representation based on end forces, enabling more efficient and accurate encoding of DLOs.
    \item We introduce a \textbf{GAT}-based dynamics model (EA-PE-GAT) for DLOs, consisting of an \textbf{E}xplicit \textbf{A}ction encoder, a \textbf{P}roperty \textbf{E}xtractor, and a graph processor. This model achieves superior prediction accuracy compared to all other existing models.
    \item We present a hybrid force-position strategy that includes state trajectory planning and MPC, enhancing both the efficiency and stability of DLO shape control.
\end{enumerate}

\section{RELATED WORK}

\subsection{DLO State Representations}

DLO state representations are categorized into four paradigms. Image-based methods use Convolutional Neural Networks (CNN) \cite{seita2021learning} or Transformer \cite{mo2022foldsformer} models to infer deformation from images but are affected by environmental factors and contain redundant information. Keypoint-based methods approximate shape by extracting keypoints, either manually \cite{yu2022global} or automatically \cite{laezza2024offline}, but lack intrinsic connections between keypoints. Graph-based methods \cite{wang2022offline} improve on this by modeling DLOs as graphs with keypoints as nodes and connections as edges, capturing structural relationships. Force \cite{bretl2014quasi} and tactile \cite{she2021cable} methods provide physical interaction information but focus on local states and require predefined physical parameters. Integration of multimodal information is still underexplored.

\subsection{DLO Dynamics Models}

For DLO dynamics modeling, analytical methods like Mass-Spring-Damper Systems (MSSs) \cite{monguzzi2025optimal} are intuitive but limited in complex elasticity, while Position-Based Dynamics (PBD) \cite{macklin2016xpbd} struggles with constraint rigidity and Finite Element Methods (FEMs) \cite{koessler2021efficient} are computationally expensive. Data-driven approaches such as Jacobian-based methods \cite{yu2022global,lagneau2020automatic} are only suitable for small deformations, whereas Graph Neural Networks (GNN) like IN \cite{battaglia2016interaction}, PropNet \cite{li2019propagation}, and GA-Net \cite{gu2025learning} have shown promise in modeling internal interactions of DLOs but still face limitations in robot action encoding and DLO property representation.

\subsection{Model-Free and Model-Based DLO Manipulation}

 Model-free methods, including reinforcement learning (RL) \cite{daniel2023multi} and imitation learning (IL) \cite{kudoh2015air}, face challenges such as low sample efficiency, high training costs, and limited generalization. Model-based methods use either action sequence shooting \cite{shi2024robocraft, huang2023learning} with MPC or state trajectory exploration \cite{sintov2020motion}, depending heavily on the model's accuracy and generalization ability. We propose a hybrid framework combining these two model-based methods for efficient and stable DLO shape control.

\section{PROBLEM FORMULATION}

\begin{figure}[htbp]
    \centering
    \includegraphics[width=0.48\textwidth]{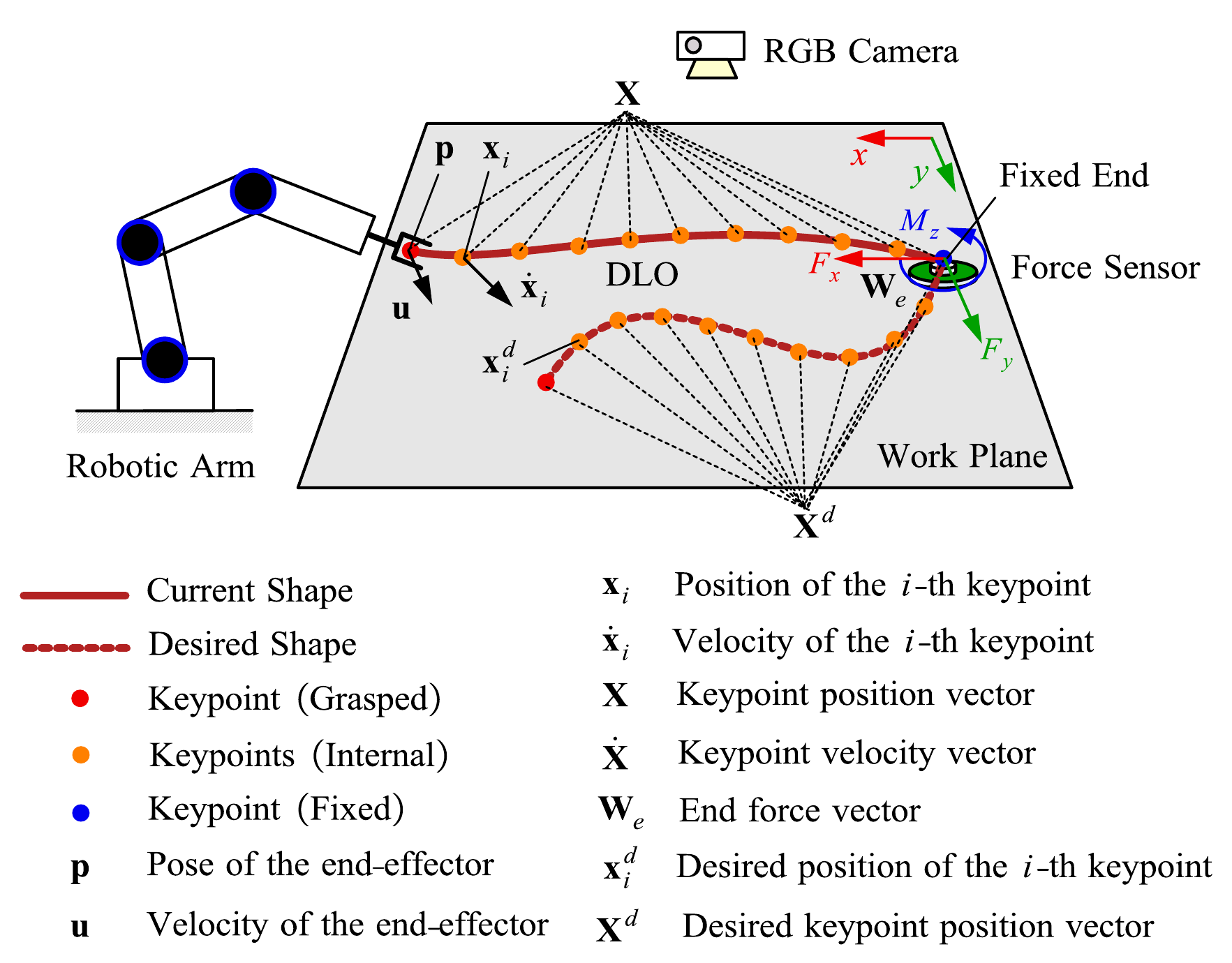}
    \caption{Setup and notations of the DLO shape control task. The robotic arm grips one end of the DLO, while the other end is fixed to a force sensor, which defines the origin of the world coordinate system (axes shown at the top right). The $\mathbf{x}$ and $\mathbf{y}$ axes lie on the work plane, and the $\mathbf{z}$ axis is perpendicular to the plane (pointing upward). }
    \label{fig:problem-formulation}
\end{figure}

This paper addresses the quasi-static shape control of elastic DLOs using both force and position information. As illustrated in Fig. \ref{fig:problem-formulation}, the robotic arm's end-effector grips one end of the DLO, while the other end is fixed to the force sensor. The force sensor is used to measure the external force exerted on the DLO. The shape of the DLO is represented by a set of evenly distributed keypoints along its length. The objective is to plan the velocity of the end-effector in order to manipulate the DLO from its current shape to the desired shape, based on the perceived visual information from the RGB camera and the force information from the force sensor.

The main definitions and notations are shown in Fig. \ref{fig:problem-formulation}. The position of the $i$-th keypoint is denoted as $\mathbf{x}_i \in \mathbb{R}^2$. The overall shape of the DLO is represented by a serious of keypoints $\mathbf{X}:=\left[ \mathbf{x}_1, \cdots, \mathbf{x}_m \right] \in \mathbb{R}^{2m}$, where $m$ is the number of the keypoints. The velocity of the $i$-th keypoint is denoted as $\dot{\mathbf{x}}_i \in \mathbb{R}^2$. The velocity of the DLO is represent by $\dot{\mathbf{X}}:=\left[ \dot{\mathbf{x}}_1, \cdots, \dot{\mathbf{x}}_m \right] \in \mathbb{R}^{2m}$. The desired position of the $i$-th keypoint is denoted as $\mathbf{x}^d_i \in \mathbb{R}^2$. The desired shape of the DLO is represented by $\mathbf{X}^d := \left[ \mathbf{x}^d_1, \cdots, \mathbf{x}^d_m \right] \in \mathbb{R}^{2m}$. We denote the force applied to the fixed end of the DLO as $\mathbf{W}_e := \left[ 
F_x, F_y, M_z \right] \in \mathbb{R}^3$, where $F_x$ is the force along the $x$-axis, $F_y$ is the force along the $y$-axis, and $M_z$ is the moment around the $z$-axis. The pose of the end-effector is denoted as $\mathbf{p} := \left[ \mathbf{r}_e, \theta_e \right] \in \mathbb{R}^3$, where $\mathbf{r}_e \in \mathbb{R}^2$ represents the position and $\theta$ represents the orientation. The velocity of the end-effector is denoted as $\mathbf{u} := \left[ d\mathbf{r}_e, d\theta_e \right] \in \mathbb{R}^3$, where $d\mathbf{r}_e \in \mathbb{R}^2$ represents the linear velocity and $d\theta$ represents the angular velocity.

\section{METHOD}

\subsection{State Representation Scheme}

Accurate and efficient state representation is fundamental for the manipulation of DLOs. In this section, we detail a hybrid representation that fuses position and force modalities to comprehensively characterize the DLO's state. The hybrid representation overcomes the limitations of traditional methods, which typically rely on a single modality. We also design two transformers, enabling seamless conversion between the position and force representations.

\subsubsection{Position Representation}

The position representation of the DLO is based on a graph structure, where keypoints along the DLO are treated as nodes, and the spatial relationships between these nodes are modeled as edges. Specifically, the keypoints are extracted either through markers placed on the DLO or via automatic detection algorithms. Each node $\mathbf{o}_i$ stores the position information $\mathbf{x}_i$ of the corresponding keypoint, interaction information $\mathbf{u}$ (i.e., robot actions), as well as attribute information such as whether the node is fixed, grasped, or internal. The $i$-th node vector is formulated as
\begin{equation}
    \mathbf{o}_i = \left[ \mathbf{x}_i, \mathbf{u}_i, \mathbf{c}_i \right]
\end{equation}
where $\mathbf{c}_i$ denotes the attribute vector of the node.
The edges of the graph are created based on the distances between nodes. A connection radius, $r_c$, is defined, and each node forms a bidirectional connection with other nodes whose distance is smaller than $r_c$. The vector form of the edge between the $i$-th node and the $j$-th node is formulated as
\begin{equation}
    \mathbf{e}_{i,j} = \left< \mathbf{x}_i, \mathbf{x}_j \right>
\end{equation}
where $i$ denotes the index of the receiving node and $j$ denotes the index of the sending node. The graph structure is defined as
\begin{equation}
    G = \left\{ O, E \right\}
\end{equation}
where $O$ and $E$ represents the sets of all nodes and edges, respectively. This graph-based representation effectively encodes the geometric configuration of the DLO and provides a structured way to model interactions between different parts of the object.

\subsubsection{Force Representation}

The force representation focuses on capturing the physical interactions between the DLO and the environment. Inspired by \cite{bretl2014quasi} \cite{chen2023quadruped}, we use the constraint force at the fixed end to represent the state of the DLO. The force representation vector is expressed as
\begin{equation}
    \mathbf{W}_e = \left[ F_x, F_y, M_z \right].
\end{equation}
This force representation significantly reduces the dimensionality of the state space while retaining critical physical information about the DLO's equilibrium configuration. As a result, it enhances the efficiency of force-based planning and control. Moreover, it complements the position representation by providing a concise and reliable description of the DLO’s state, particularly when visual information is limited or unreliable.

\subsubsection{Representation Transformers}

To enable bidirectional conversion between position and force representations, we introduce two MLP-based transformers: the Position-to-Force Transformer (P2FT) and the Force-to-Position Transformer (F2PT). These modules map between node positions $\mathbf{X}$ and force vector $\mathbf{W}_e$, enabling flexible state transformations while maintaining low computational overhead.

\subsection{Dynamics Model Learning}

\begin{figure*}[htbp]
    \centering
    \includegraphics[width=0.85\textwidth]{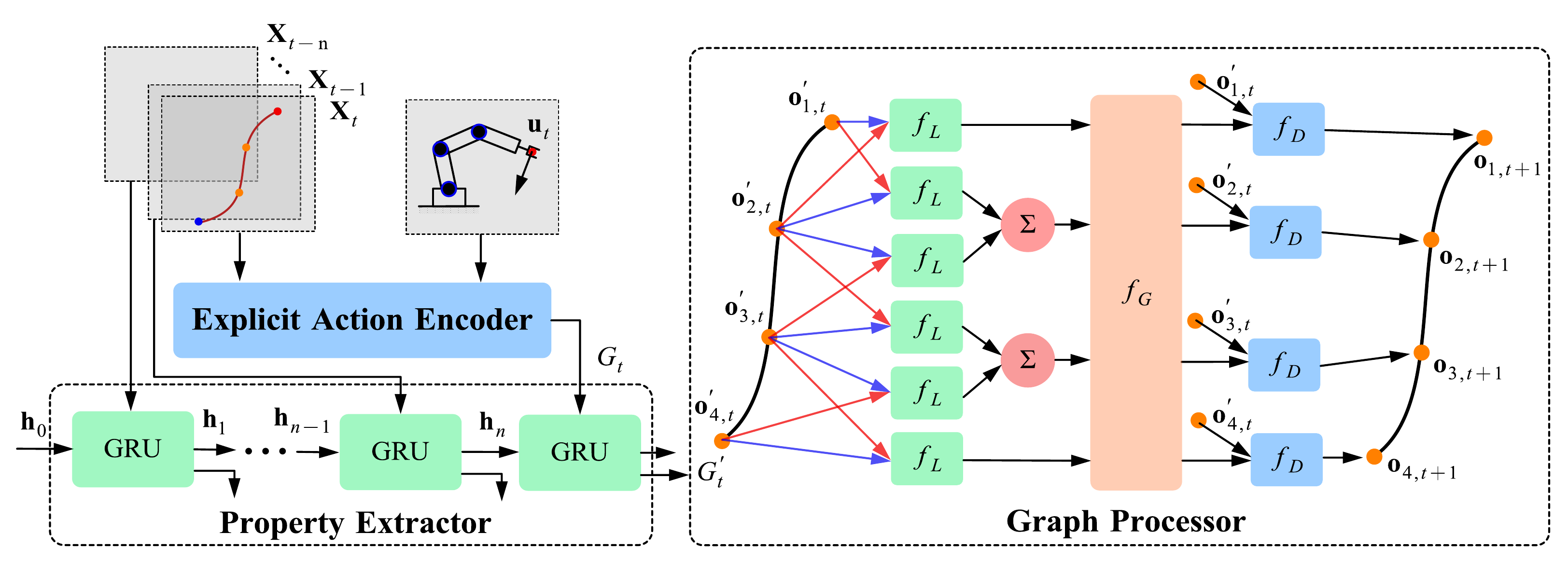}
    \caption{Schema of the EA-PE-GAT model. First, the explicit action encoder embeds the robot's actions into the graph representation of the DLO. Second, the property extractor captures the physical properties of the DLO from the historical states and output a latent graph combining the current graph and the extracted properties. Finally, the graph processor simulates the interactions between nodes in the latent graph and updates the state of the DLO.}
    \label{fig:dynamics_model}
    \vspace{-10pt}
\end{figure*}

In this section, our goal is to learn a dynamics model for predicting the future state of a DLO, given its current and historical states and the robot's actions. The proposed dynamics model (EA-PE-GAT) integrates an explicit action encoder, a property extractor, and a graph processor, as shown in \ref{fig:dynamics_model}.

\subsubsection{Explicit Action Encoder}

The explicit action encoder is designed to integrate the robot's actions into the graph representation of the DLO. This encoder assumes that any small local region of the DLO can be approximated as a rigid body. Specifically, the neighbors of the grasped node undergo rigid body motion along with the grasped node. The position update of these neighboring nodes is calculated as follows:
\begin{equation}
\label{equation:explicit-encoding-formula}
\mathbf{x}_{i}^{t+1}=\mathbf{r}_{e}^{t}+d \mathbf{r}_{e}^{t}+\left[\begin{array}{cc}
\cos d \theta_{e}^t & -\sin d \theta_{e}^t \\
\sin d \theta_{e}^t & \cos d \theta_{e}^t
\end{array}\right]\left(\mathbf{x}_{i}^{t}-\mathbf{r}_{e}^{t}\right)
\end{equation}
where $i$ is the index of the neighboring node. The explicit encoding method effectively fuses the robot's actions with the DLO's state, providing a clear physical interpretation.

\subsubsection{Property Extractor}

The property extractor is designed to automatically infer the physical properties of the nodes from the historical states of the DLO. It is based on the Gated Recurrent Unit (GRU) architecture. The input to the property extractor is a sequence of the DLO's historical states, and each state is processed successively through the GRU's reset and update mechanisms. The hidden state $\mathbf{h}_i$ of the GRU is updated iteratively, capturing the temporal dynamics of the DLO. The final output of the property extractor is a latent graph $G_t^\prime$ that integrates the current state $G_t$ with the extracted physical properties $\mathbf{h}_n$. The latent graph $G_t^\prime$ contains enriched information about the DLO's physical attributes, such as node types (fixed, internal, or grasped) and their velocities, which are very helpful for accurate dynamics prediction.

\subsubsection{Graph Processor}

The graph processor is based on the GAT and is responsible for capturing the interactions between nodes in the DLO. First, for each node, the local interactions with its neighbors are computed using a local interaction function $f_L(\cdot)$:
\begin{equation}
    \mathbf{e}^\prime_{ij,t} = f_L(\mathbf{o}^\prime_{i,t}, \mathbf{o}^\prime_{j,t})
\end{equation}
where $\mathbf{o}^\prime_{i,t}$ and $\mathbf{o}^\prime_{j,t}$ are the states of the receiving node $i$ and the sending node $j$, respectively. Second, the local interactions are aggregated using a global interaction function $f_G$ with an attention mechanism, which allows the model to weigh the importance of different local interactions:
\begin{equation}
    \mathbf{e}^\prime_{\mathrm{g},i,t} = f_G(\mathbf{E}^\prime_{\mathrm{agg,i,t}})
\end{equation}
where $\mathbf{E}^\prime_{\mathrm{agg,i,t}}$ is the aggregated local interactions received by node $i$, and $f_G(\cdot)$ is implemented using a Transformer encoder with multi-head self-attention. Finally, the state of each node is updated based on the global interaction and the node's previous state:
\begin{equation}
    \mathbf{o}_{i,t+1} = f_D(\mathbf{o}^\prime_{i,t}, \mathbf{e}^\prime_{g,i,t})
\end{equation}
where $f_D(\cdot)$ is the decoder that maps the node's state from the latent space back to the Euclidean space, providing the predicted position of the node. The local interaction function $f_L(\cdot)$ and the decoder $f_D(\cdot)$ both adopt the MLP structure and are shared by all nodes and edges in the graph to reduce the number of model parameters.

\subsection{Hybrid Force-Position Strategy}

This section introduces the proposed hybrid force-position strategy for shape control of DLOs, as illustrated in Fig. \ref{fig:overview-strategy}. The whole framework consists of two main components: state trajectory planning and MPC.

\subsubsection{State Trajectory Planning}

The goal of state trajectory planning is to generate a smooth and feasible trajectory that guides the DLO from its initial shape to the target shape. This is achieved through the following steps:
\begin{description}[leftmargin=1em]
  \item[Step 1:] Target shape transformation. The target shape of the DLO, represented in position space, is converted to the force representation using the P2FT.
  \item[Step 2:] Force trajectory planning. With the initial and target force representations, a series of intermediate force waypoints are generated using linear interpolation.
  \item[Step 3:] Conversion to position space. Each intermediate force waypoint is then converted back to the position space using the F2PT, and the position waypoints serve as sub-goals for the MPC controller.
\end{description}

\subsubsection{Model Predictive Control}

MPC is employed to execute the planned trajectory and achieve shape control of the DLO. The control problem is formulated as
\begin{equation}
    \begin{aligned}
        \min_{\hat{\mathbf{u}}_{0:T-1}} & \quad \left\| \hat{\mathbf{X}}_{T}-\mathbf{X}^{d}\right\| _{2}^{2} \\
        \text{s.t.} & \quad \hat{\mathbf{X}}_{t + 1}=\phi\left(\hat{\mathbf{X}}_{t},\hat{\mathbf{u}}_{t}\right) \\
        & \quad \left\| \hat{\mathbf{u}}_{t}\right\| \leq\left\| \mathbf{u}_{\max }\right\|
    \end{aligned}
\end{equation}
where $T$ is the total number of time steps, $\phi(\cdot)$ is the dynamics model of the DLO, and $\mathbf{u}_{\max}$ is the maximum allowable control input for the robot.
The implementation of the MPC is summarized in Algorithm \ref{alg:MPC}. The MPC controller utilizes EA-PE-GAT to accurately predict the DLO's future state. At each time step, the MPC controller optimize the control sequence by minimizing the difference between the predicted and target shapes. We use the $L_2$ loss to calculate the shape error and select stochastic gradient descent (SGD) as the optimizer. Only the first step of the optimized control sequence is executed, and the process repeats at the next time step. This rolling-optimization mechanism allows the controller to adapt to real-time changes in the DLO's state and ensures robustness against uncertainties and disturbances.

\begin{algorithm}
  \caption{Model Predictive Control}
  \label{alg:MPC}
  \small
  \begin{algorithmic}[1]
    \REQUIRE Dynamics model $\phi(\cdot)$, Target shape $\mathbf{X}^d$, Current control inputs $\hat{\mathbf{u}}_{0:T-1}$, Time horizon $T$, Number of optimization steps $N$, Learning rate $\eta$
    \FOR{$t = 0, \cdots, T-1$}
        \STATE $\mathbf{X}_t \leftarrow \mathrm{getCurrentState()}$
        \STATE $\hat{G_t} \leftarrow \mathrm{constructGraph(\mathbf{X}_t)}$
        \FOR{$i = t, \cdots, T-1$}
            \STATE $\hat{G}_{i+1} \leftarrow \phi(\hat{G}_i, \hat{\mathbf{u}}_i)$
        \ENDFOR
        \FOR{$n = 1 \cdots, N$}
            \STATE $\hat{\mathbf{u}}_{t:T} \leftarrow \hat{\mathbf{u}}_{t:T} - \eta \nabla_{\hat{\mathbf{u}}_{t:T}}L(\hat{\mathbf{X}}_T,\mathbf{X}^d)$
        \ENDFOR
        \STATE $\mathrm{executeAction}(\hat{\mathbf{u}}_t)$
    \ENDFOR
  \end{algorithmic}
\end{algorithm}

\section{EVALUATION}

\subsection{Simulation and Real-world Experiment Setup}

\subsubsection{Simulation Setup}

For training and validation, we develop a simulation environment based on MuJoCo \cite{todorov2012mujoco}. The DLO is composed of 40 articulated capsules, with a total length of 1 m, a diameter of 5 mm, a bending stiffness of $10^7$, and a damping coefficient of 0.05. We uniformly selected 11 keypoints along the DLO. One end of the DLO is fixed to a force sensor for data collection, while the other end is rigidly connected to the end-effector.

\subsubsection{Real-World Experiment Setup}

\begin{figure}[htbp]
    \centering
    \includegraphics[width=0.48\textwidth]{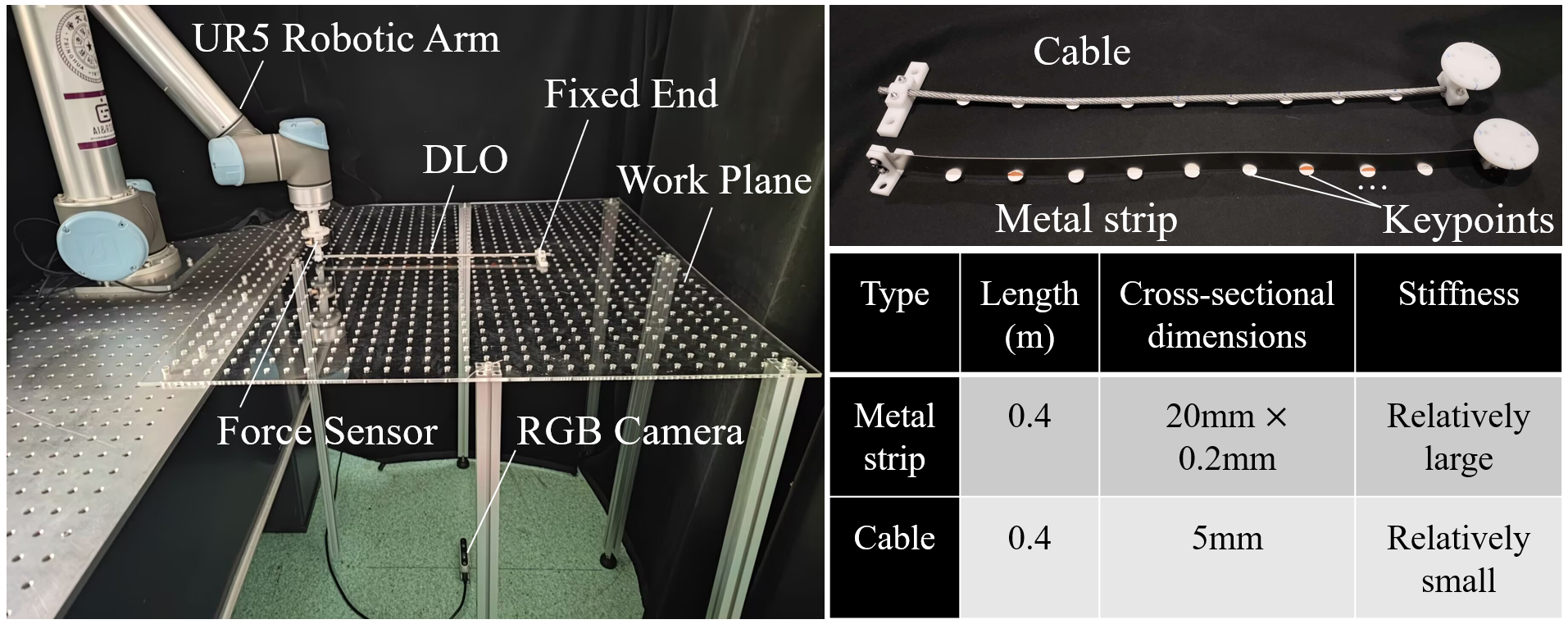}
    \caption{Global view of the real-world experiment setup and DLOs used in the experiment.}
    \label{fig:experiment_setup}
\end{figure}

The real-world experiment setup is shown in Fig. \ref{fig:experiment_setup}. A transparent acrylic plate is used as the works plane, with a UR5 robotic arm mounted on one side for manipulating the DLO. An RGB camera is installed directly beneath the work plane to avoid occlusions. To achieve accurate and robust detection of DLO keypoints, we additionally train a customized semantic segmentation model based on a Feature Pyramid Network (FPN) architecture. This model uses a ResNet-34 backbone pretrained on ImageNet to extract multi-scale features and is fine-tuned for keypoint-level binary segmentation. A force sensor is attached to the UR5's flange to measure the constraint force at the DLO's fixed end indirectly. Two types of DLOs, a metal strip and a cable, are used in the experiments, each marked with 11 keypoints (including the two endpoints), following the setting used in prior work \cite{huang2023learning}.

\subsection{Data Collection and Training Details}

We adopt the data collection scheme in \cite{gu2025learning}, where the end-effector moves towards the destination randomly sampled in the workspace to continually generate data, including the robot actions, the positions of the keypoints, and the force data at each time step. A total of 3000 trajectories are collected in the simulation, with each trajectory containing 150 time steps and a time interval of 0.1 seconds. In the real world, we collect 1000 trajectories for both the metal strip and the cable, with each trajectory containing 100 time steps and a time interval of 0.2 seconds.

The dataset is split into a training set, a validation set and a test set at a ratio of 8:1:1. The training settings for the representation transformers and the dynamics model are the same. The training batch size is set to 128. The Adam optimizer is chosen, and the learning rate decays from $10^{-3}$ to $10^{-6}$. The $L_2$ loss function is used to calculate the one-step prediction error. All models are run on an NVIDIA RTX 3090 GPU.

 The P2FT consists of 5 hidden layers with dimensions of 64, 128, 128, 64, and 16, respectively. The F2PT consists of five hidden layers with dimensions of 16, 64, 128, 128, and 64, respectively. The property extractor adopts 2 stacks of GRU with a dimension of 128. The Transformer encoder in the graph processor has a dimension of 128, with 6 layers and 12 heads, and the dimension of the feed-forward network is 512. The local interaction function $f_L(\cdot)$ and the decoder $f_D(\cdot)$ both consist of three 128-dimensional hidden layers. MLP and GRU modules are initialized with zero biases and normally distributed.

\subsection{Simulation Results}

\subsubsection{Representation Transformers}

\begin{figure}[htbp]
    \centering
    \includegraphics[width=0.48\textwidth]{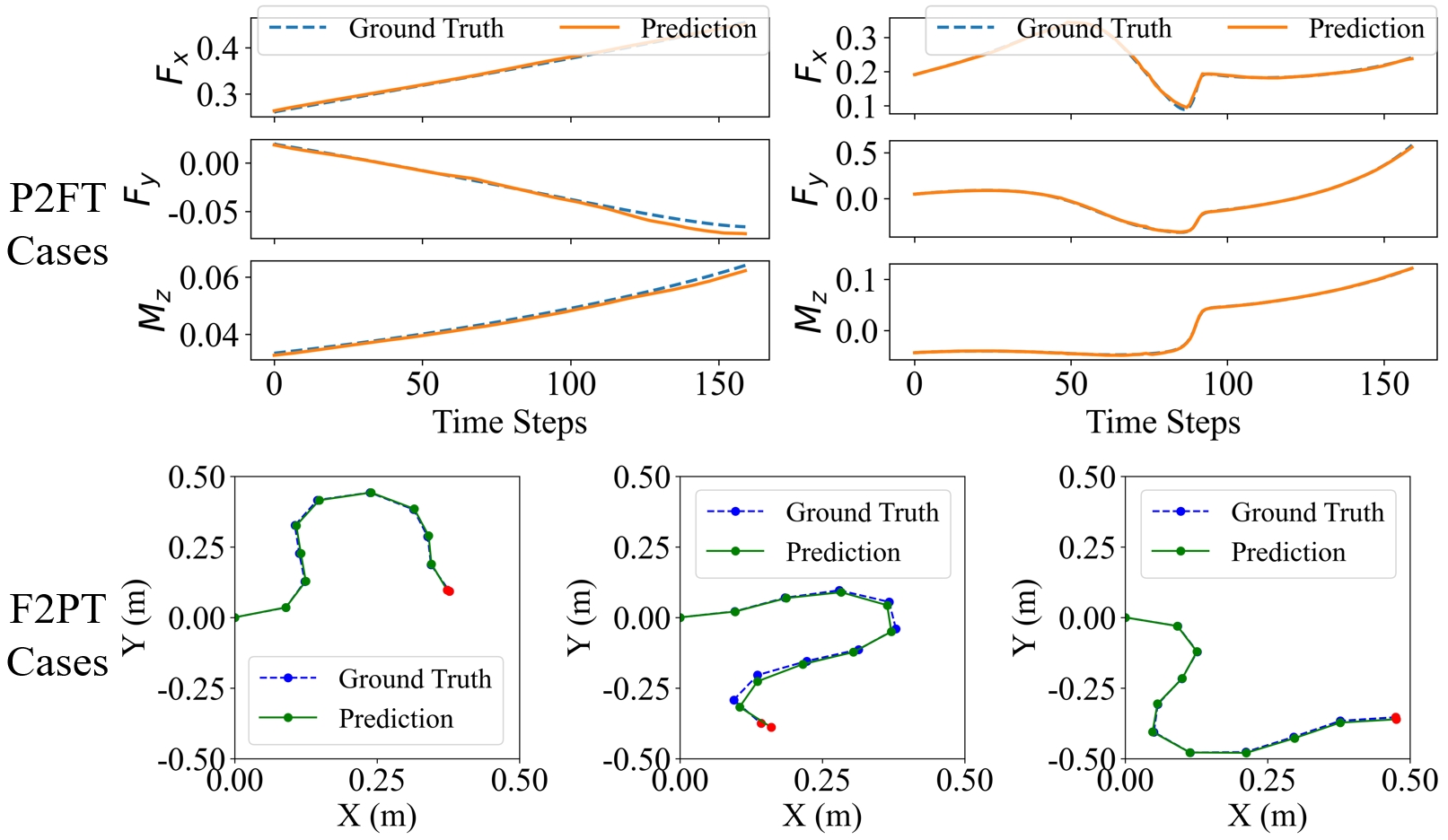}
    \caption{Cases of the prediction results of the P2FT and F2PT.}
    \label{fig:results_transformers}
\end{figure}

The P2FT achieves an average prediction error of [$ 1.73\times10^{-3}$,\,$1.91\times 10^{-3}$,\,$3.52\times10^{-4}$] for the three components of the force representation. The F2PT achieves an average prediction error of 9.72 mm. Some prediction cases are shown in Fig. \ref{fig:results_transformers}, demonstrating high accuracy of the transformers. The inference times of the P2FT and F2PT are 0.44 ms and 0.43 ms, respectively, showing excellent real-time performance.

We also explored more complex architectures for state transformation. For P2FT, we implement a Transformer-based architecture similar to ViT, where each keypoints are treated as tokens and the [CLS] token are used to regress the force vector. For F2PT, we adopt a DiT-style denoising transformer, conditioning on the target force and generating keypoint positions. However, these models yielded higher prediction errors [$ 1.95\times10^{-3}$,\,$2.45\times 10^{-3}$,\,$4.10\times10^{-4}$] and 52.3mm, respectively, along with longer inference times 2.94 ms and 227 ms. This suggests that the transformation task is sufficiently low-dimensional, and MLPs offer a better trade-off between accuracy and efficiency.

\subsubsection{Dynamics Model}

We compare the EA-PE-GAT with four baselines:
\begin{itemize}
    \item MLP: The model consists of five hidden layers with dimensions of 64, 128, 128, 128, and 64, respectively.
    \item GA-Net: The model, proposed in \cite{gu2025learning}, adopts the GAT architecture with an MLP-based graph encoder.
    \item EA-GAT: The model replaces the property extractor in EA-PE-GAT with an MLP that has three 128-dimensional hidden layers.
    \item PE-GAT: The model removes the explicit action encoder from EA-PE-GAT, directly concatenating the action vector with object state vector to form an action-state vector. The action-state vector sequence is then fed into the property extractor to be encoded into a latent graph.
\end{itemize}

\begin{figure}[htbp]
    \centering
    \includegraphics[width=0.43\textwidth]{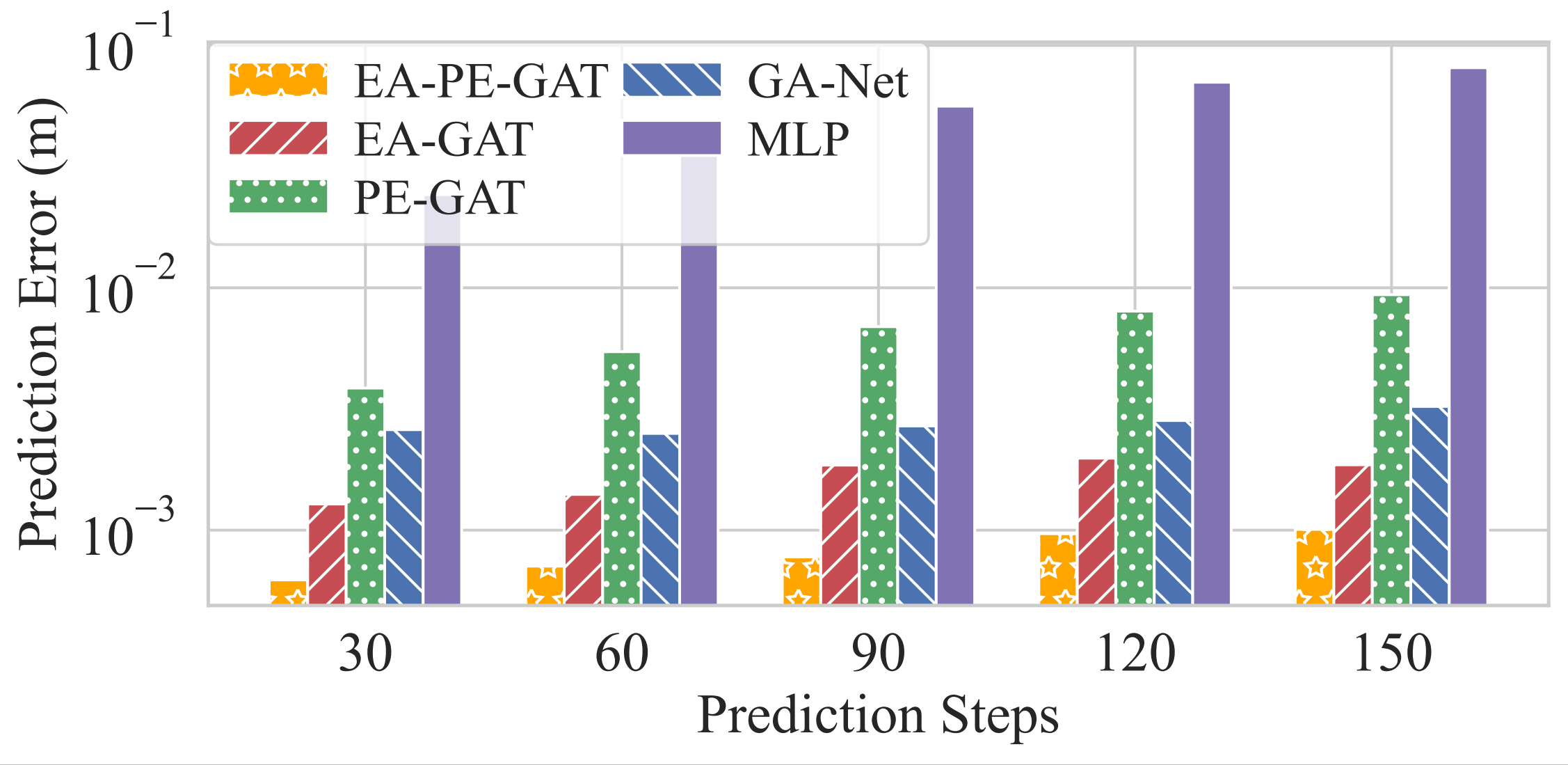}
    \caption{Multi-step prediction errors of the dynamics models. The prediction error is defined as the average distance between the predicted positions and actual positions of the keypoints.}
    \label{fig:GNN_error}
\end{figure}

We perform multi-step prediction tests on the dynamics models, and the results are shown in Fig. \ref{fig:GNN_error}. Root Mean Squared Error (RMSE) is used to measure the difference between the predicted states and the ground-truth states of keypoints. As illustrated in Fig. \ref{fig:GNN_error}, EA-PE-GAT exhibits consistently superior performance across all prediction steps. The comparative results are discussed below.

The GAT-based models achieve significantly lower prediction errors compared to the MLP model, indicating the GAT architecture is more suitable for modeling the dynamics of DLOs. The GAT captures the spatial relationships and interactions between the keypoints more effectively than the MLP model. EA-GAT, with an explicit action encoder, shows a significant improvement in prediction accuracy, reducing errors by approximately 40\% compared to GA-Net. The result suggests that the explicit action encoder allows for a more precise and meaningful integration of the robot's action into the dynamics prediction. By encoding the actions explicitly, the model can better understand the influence of the robot's action on the DLO's state.

EA-PE-GAT, with the addition of the property extractor, improves significantly in performance, reducing prediction errors by about 50\% compared to EA-GAT. The result reveals that the property extractor efficiently captures highly representative attribute features of the DLO from historical states. These features help enhance the model's expressive power, enabling it to make more precise predictions in complex dynamic systems. In contrast, PE-GAT shows a higher RMSE than GA-Net. The reason could be that the absence of the explicit action encoder increases the learning difficulty of the property extractor, which also highlights the necessity of the explicit action encoder.

Additionally, we notice that prediction errors do not increase monotonically with prediction steps in some models (like GA-Net and EA-GAT). This phenomenon may be attributed to the injection of 5\% noise in the training data, which helps the models learn to handle uncertainties and avoid overfitting.

\begin{figure}[htbp]
    \centering
    \includegraphics[width=0.43\textwidth]{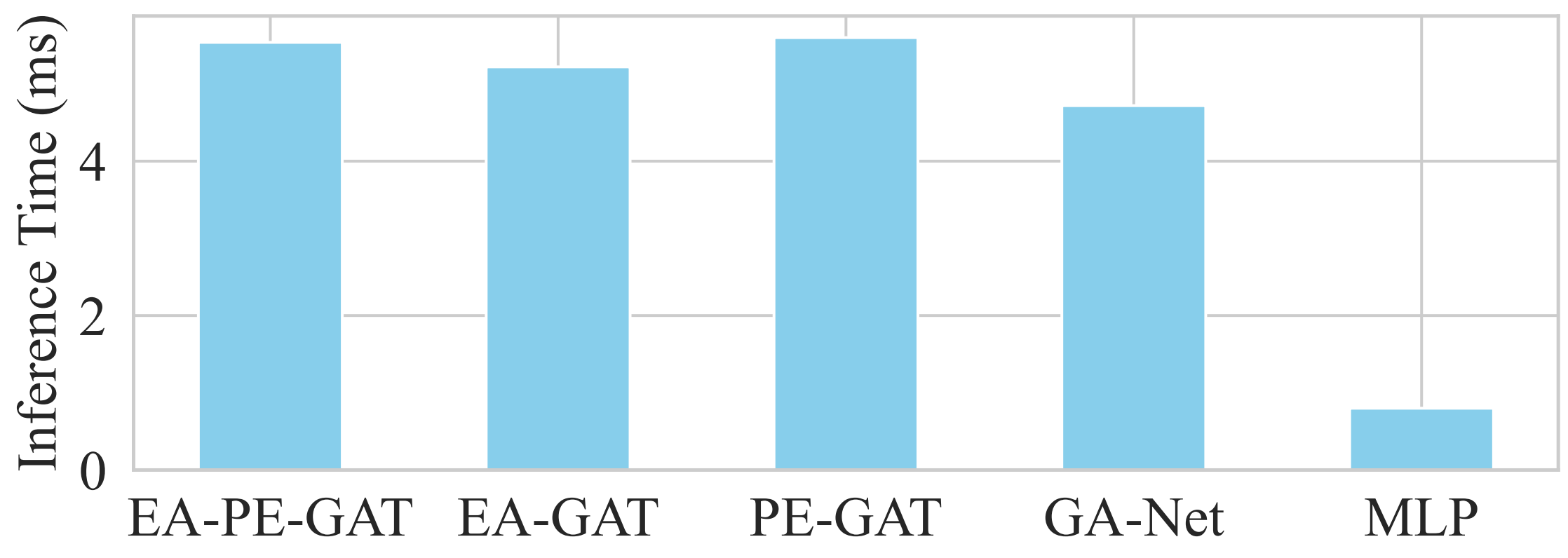}
    \caption{Single-step inference time of the dynamics models.}
    \label{fig:GNN_time}
\end{figure}

The statistics of single-step inference time for the dynamics models are shown in Fig. \ref{fig:GNN_time}. Although GAT-based models have a higher computational cost than MLP models, they achieve significantly higher prediction accuracy, with errors one or two orders of magnitude lower. Given that inference time is not the primary optimization target, GAT-based models, which achieve high-precision predictions within acceptable inference time, are the better choice. Furthermore, comparing various GAT-based models reveals that the explicit action encoder and property extractor consume very little time but significantly improve prediction accuracy.

\subsubsection{Shape Control Strategy}

We compare the proposed hybrid force-position strategy (denoted as Hybrid) with two baselines:
\begin{itemize}
    \item Position-based MPC (denoted as P-MPC): This framework uses the same MPC controller as the hybrid force-position strategy but without state trajectory planning. It directly uses the position representation of the target shape to compute the cost function.
    \item Model-Free Reinforcement Learning (denoted as RL): This framework employs the Proximal Policy Optimization (PPO) algorithm. The observation state is the keypoint positions, and the action is the three-dimensional movement of the end-effector. The reward function is set as the negative RMSE between the current and target shapes. The actor and critic policy networks both use an MLP with three 512-dimension hidden layers.
\end{itemize}

The experiments include both large and small deformation tasks. Eight pairs of significantly different shapes (large deformation) and two pairs of slightly different shapes (small deformation) are randomly selected as initial and target shapes for 10 tasks. The evaluation metric is the RMSE between the current and target keypoint positions. The hybrid force-position strategy is configured with 4 waypoints, and the time step limit to reach each waypoints is set to 50. The error threshold for switching waypoints is 0.03 m. For the position-based MPC, the prediction horizon is set to 200 time steps. The RL has a maximum execution step limit of 200. The maximum velocity of the robot's action is set to $\left[ \mathrm{0.05 \, m/s, 0.05 \, m/s, 0.2 \, rad/s} \right]$. The error threshold for success is set to 0.01 m.

The statistical results of the simulation experiments are presented in Table \ref{tab:sim_hybrid_results}. The hybrid force-position strategy achieved significantly lower shape errors compared to Position-based MPC and RL. This is because the hybrid force-position strategy successfully completed all tasks, while Position-based MPC and RL struggled with large deformation tasks, achieving success rates of only 12.5\% and 25\%, respectively.

\begin{table}[htbp]
    \caption{Success Rate and Shape Error of Manipulation Frameworks in the Simulation}
    \label{tab:sim_hybrid_results}
    \centering
    \begin{tabular}{lccc}
        \toprule
        \multirow{2}{*}{Method} & \multicolumn{2}{c}{Success Rate $\uparrow$}    & \multirow{2}{*}{RMSE (mm)$\downarrow$} \\ \cmidrule{2-3}
                                & Large Deformation          & Small Deformation           &                       \\ \midrule
                               P-MPC  & 12.5\% & 100\% & 189                \\
                               RL & 25\% & 100\% & 94.6                \\ 
                               \textbf{Hybrid} & \textbf{100\%} & \textbf{100\%} & \textbf{8.44} \\
                               \bottomrule
    \end{tabular}
\end{table}

\begin{figure}[htbp]
    \centering
    \includegraphics[width=0.48\textwidth]{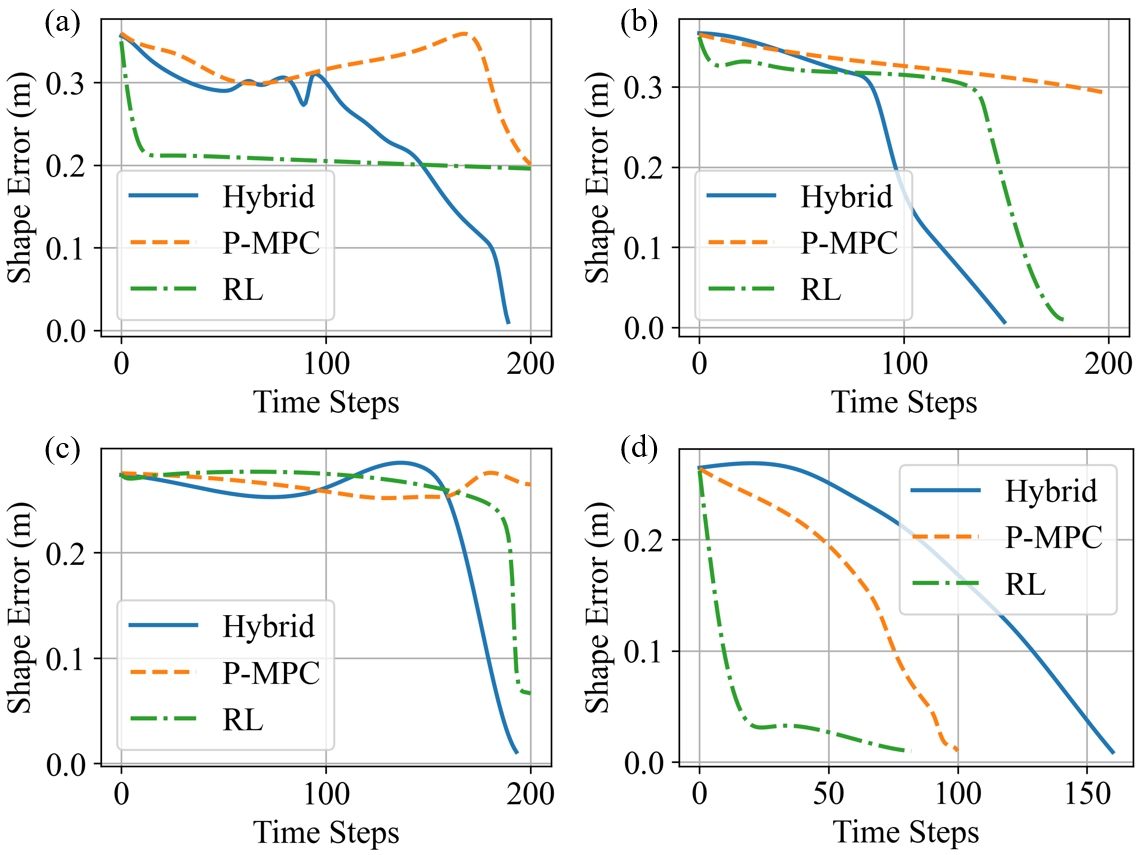}
    \caption{Shape error variation curves of the manipulation frameworks in four tasks. (a)-(c) Large deformation tasks. (d) Small deformation task.}
    \label{fig:shape_error}
\end{figure}

\begin{figure}[htbp]
    \centering
    \includegraphics[width=0.48\textwidth]{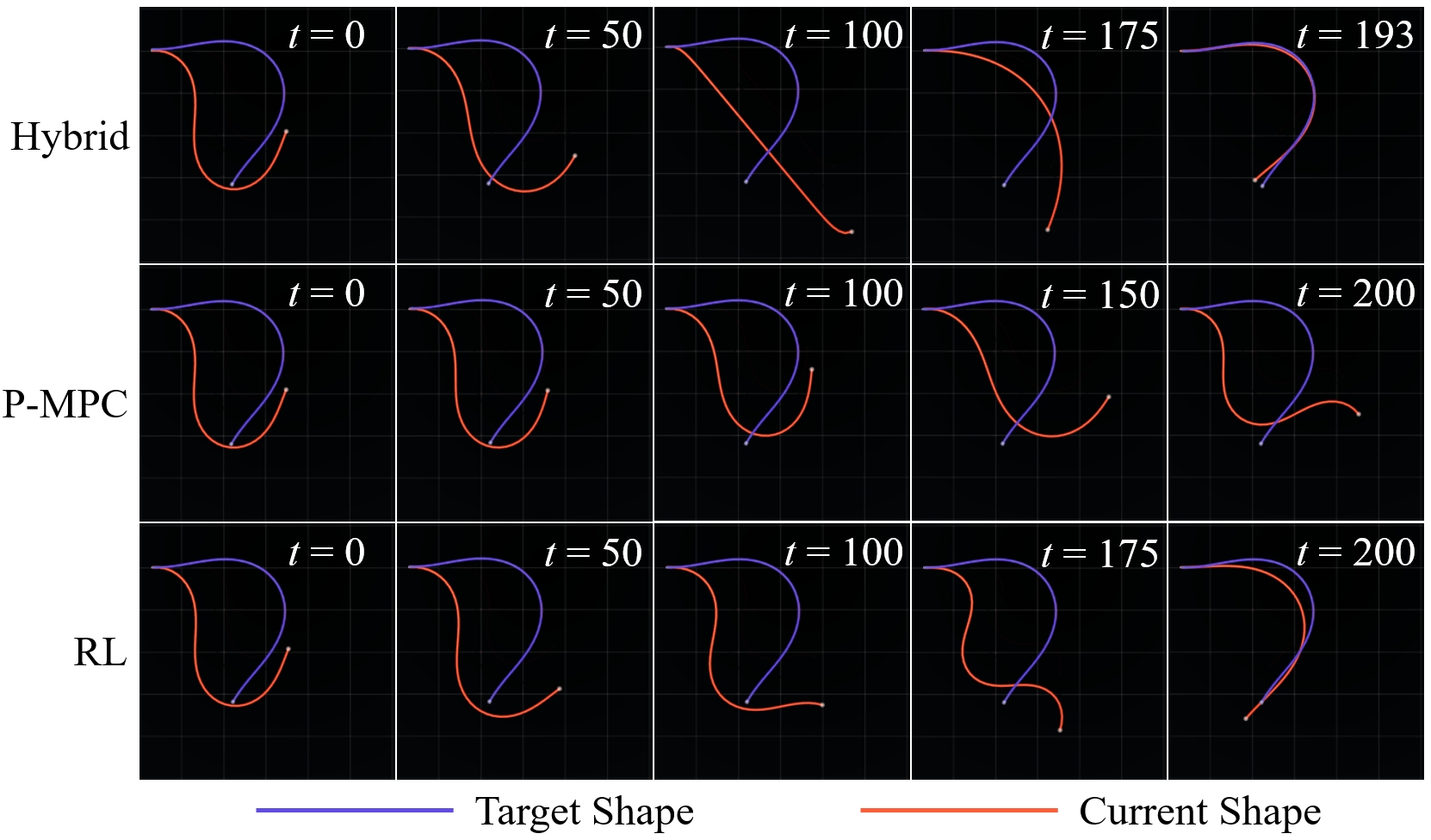}
    \caption{DLO shape control processes of the manipulation frameworks in the large deformation task.}
    \label{fig:sim_task}
\end{figure}

The shape error variation curves for four tasks (including three large deformation tasks and one small deformation task) are depicted in \ref{fig:shape_error}. Among the evaluated methods, the hybrid force-position strategy demonstrates the highest efficiency, completing large deformation tasks in the least number of time steps. In contrast, Position-based MPC and RL exhibits slow error reduction in most cases, failing to complete the tasks within the specified time steps. This indicates that relying solely on position representation is prone to local optima, significantly reducing the efficiency of manipulation. Although RL has some ability to escape local optima through random exploration, its efficiency is still greatly affected. The hybrid force-position strategy, by planning in the force space and converting the waypoints to position space, effectively decomposed the large deformation task into multiple small deformation tasks, avoiding local optima. To provide a more intuitive comparison, the execution process of the three frameworks in the third task (corresponding to Fig. \ref{fig:shape_error} (c)) is shown in Fig. \ref{fig:sim_task}.

The hybrid force-position strategy demonstrates reduced efficiency in small deformation scenarios. In such cases, there is typically no risk of falling into local optima, and the force-space planning mechanism may result in unnecessary "detour", leading to wasted time steps. This suggests the need for an adaptive waypoint selection mechanism, where the number and spacing of intermediate targets are dynamically adjusted based on the initial shape error or the distance in force space, which will be considered as our future work. 

\begin{figure}[htbp]
    \centering
    \includegraphics[width=0.17\textwidth]{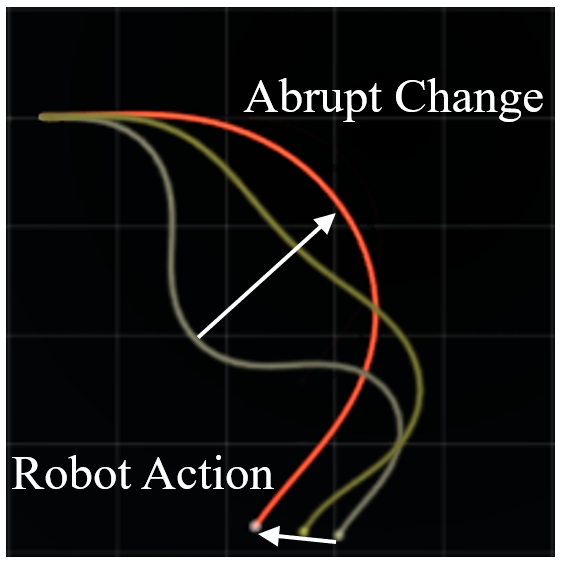}
    \caption{Illustration of the abrupt shape change occurred in the manipulation process of RL. A small robot action triggers a large object deformation.}
    \label{fig:abrupt_change}
\end{figure}

As illustrated in Fig. \ref{fig:sim_task}, in the 175th to 200th time steps of RL, the DLO exhibits an abrupt transition in shape (from an "M" configuration to an upward convex form) despite minimal movement of the end-effector. This phenomenon is further visualized in Fig. \ref{fig:abrupt_change} and can be attributed to the underactuated nature of the system, where limited control inputs are insufficient to fully govern the DLO’s infinite degrees of freedom. As a result, the DLO spontaneously transitions from high-energy to low-energy configurations, leading to sharp fluctuations in shape error. These sudden deformations challenge the responsiveness of the controller and may cause temporary loss of control. In safety critical scenarios, such as obstacle avoidance, this behavior could increase the risk of unintended collisions. At the 200th time step of the Position-based MPC in Fig. \ref{fig:sim_task},  the DLO exhibits a similar “M” configuration, indicating a potential risk of abrupt change. In contrast, the hybrid force-position strategy achieves smooth and stable shape transitions in all tasks.

\subsection{Real-World Results}

\begin{figure}[htbp]
    \centering
    \includegraphics[width=0.48\textwidth]{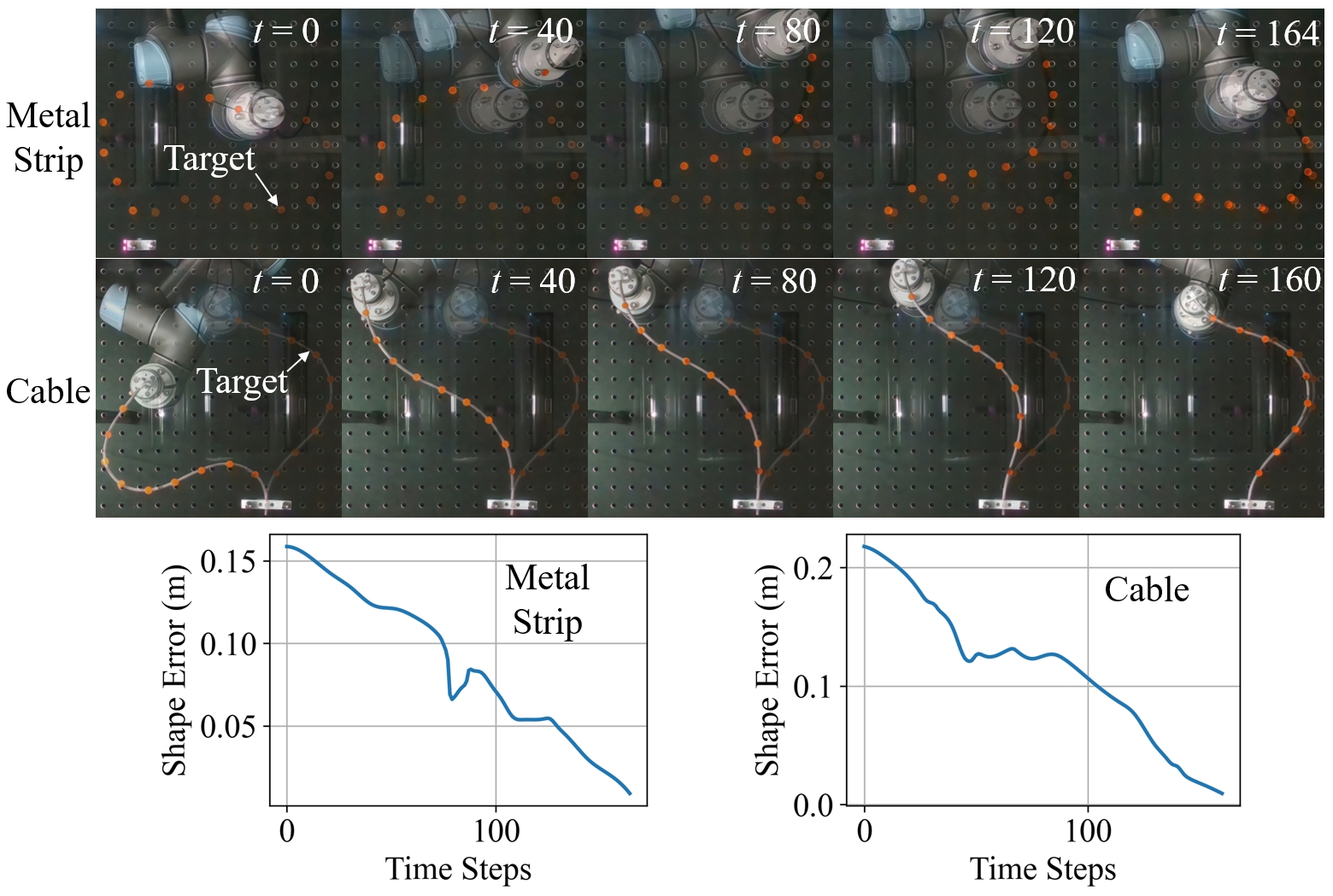}
    \caption{Cases of real-world experiments for the hybrid force-position strategy using the metal strip and cable.}
    \label{fig:exp_task}
\end{figure}

We validate the hybrid force-position strategy in the real world. All models are retrained using data collected from the real world and demonstrate consistent performance with that in the simulation. Since the metal strip and cable are shorter than the DLO in the simulation, we adjust the maximum speed of the end-effector to $\left[ \mathrm{0.02 \, m/s, 0.02 \, m/s, 0.2 \, rad/s} \right]$ and set the error threshold for waypoint switching to 0.02 m, while keeping the other parameters the same as in the simulation. We conduct four large deformation experiments for both the metal strip and cable, and one case for each type is shown in Fig. \ref{fig:exp_task}. The hybrid force-position strategy is able to complete all tasks within the specified number of steps without any abrupt shape change of the DLO during the manipulation, demonstrating its efficiency and stability.

\section{CONCLUSION}

In this study, we propose a hybrid force-position strategy for the shape control of DLOs using a combination of force and position information. The proposed method includes a novel state representation scheme combining position and force representations, a GAT-based dynamics model (EA-PE-GAT), and a manipulation framework integrating state trajectory planning and MPC. Extensive simulation and real-world experiments demonstrate the superiority of the proposed dynamics model and the effectiveness of the hybrid force-position strategy. Our approach successfully decomposes large deformation tasks into multiple small deformation tasks, avoiding local optima and ensuring smooth transitions, thereby improving the efficiency and stability of DLO shape control. This research advances DLO manipulation and offers insights into robotic manipulation strategies with multimodal state representations. Future work will focus on refining the state trajectory planning for small deformation tasks to reduce unnecessary time steps and extending the framework to three-dimensional space.

\addtolength{\textheight}{-12cm}   






\bibliographystyle{ieeetr}  
\bibliography{main}     




\end{document}